\documentclass[12pt]{article}

\usepackage{amssymb,amsmath,amsfonts,latexsym,graphicx}
\usepackage[numbers]{natbib}
\usepackage{hyperref}

\begin{document}















\begin{center}
{\LARGE \textbf{Automated Bleeding Detection and\\
Classification in Wireless Capsule Endoscopy\\
with YOLOv8-X}}

\vspace{0.5cm}

Pavan C Shekar$^{a}$, Vivek Kanhangad$^{b}$, Shishir Maheshwari$^{c}$, T Sunil Kumar$^{d}$

\vspace{0.5cm}

\small
$^{a}$ Indian Institute of Technology, Indore (IITI), pavancshekar60@gmail.com\\[0.2em]
$^{b}$ Indian Institute of Technology, Indore (IITI), kvivek@iiti.ac.in\\[0.2em]
$^{c}$ Thapar Institute of Engineering \& Technology (TIET), shishir.maheshwari@thapar.edu\\[0.2em]
$^{d}$ Vellore Institute of Technology, Chennai (VIT), suneel457.ece@gmail.com
\end{center}

\vspace{0.5cm}

\begin{abstract}
Gastrointestinal (GI) bleeding, a critical indicator of digestive system disorders, requires efficient and accurate detection methods. This paper presents our solution to the Auto-WCEBleedGen Version V1 Challenge, where we achieved the consolation position. We developed a unified YOLOv8-X model for both detection and classification of bleeding regions in Wireless Capsule Endoscopy (WCE) images. Our approach achieved 96.10\% classification accuracy and 76.8\% mean Average Precision (mAP) at 0.5 IoU on the validation dataset. Through careful dataset curation and annotation, we assembled and trained on 6,345 diverse images to ensure robust model performance. Our implementation code and trained models are publicly available at \url{https://github.com/pavan98765/Auto-WCEBleedGen}.
\end{abstract}

\section{Introduction}\label{sec1}

Gastrointestinal (GI) bleeding represents a critical medical condition affecting approximately 150 per 100,000 adults annually, with mortality rates ranging from 5\% to 30\% depending on the severity and location of bleeding. This condition, characterized by blood loss anywhere in the digestive tract from the esophagus to the rectum, manifests through various symptoms including hematemesis (vomiting of blood), melena (black, tarry stools), or hematochezia (bright red blood in stools). The diverse etiology encompasses multiple disorders, from peptic ulcers and inflammatory bowel disease to malignancies, necessitating rapid and accurate diagnosis for optimal patient outcomes.

Traditional diagnostic approaches for GI bleeding have relied heavily on conventional endoscopy, which, while effective, can be invasive and may not reach certain portions of the small intestine. The advent of Wireless Capsule Endoscopy (WCE)\cite{survey_paper} in 2000 revolutionized gastrointestinal imaging by offering a non-invasive method to visualize the entire GI tract, particularly the small bowel, which was previously challenging to examine. This innovative technology consists of a swallowable capsule containing a miniature camera that captures thousands of images as it travels through the digestive system, providing detailed visualization of the mucosa and potential bleeding sites.

However, the clinical implementation of WCE faces significant practical challenges in data analysis. A single examination typically generates between 60,000 to 100,000 video frames, requiring healthcare professionals to spend approximately 2-3 hours reviewing the footage. This time-intensive process is not only resource-demanding but also susceptible to human error due to fatigue and the monotonous nature of the task. Moreover, the real-time detection of bleeding regions, which could be crucial in emergency situations, remains particularly challenging due to the vast amount of data that needs to be processed quickly and accurately.

The Auto-WCEBleedGen Challenge emerged as a response to these limitations, aiming to catalyze the development of automated methods for bleeding detection and classification in WCE images. The challenge seeks to leverage advanced computational techniques to assist healthcare providers in analyzing WCE data more efficiently and accurately.

Our solution to this challenge employs a unified deep learning approach, specifically designed to streamline this critical medical imaging task while maintaining high diagnostic accuracy. In this study, we present our comprehensive solution that not only addresses the immediate need for automated bleeding detection but also considers the broader clinical context in which these tools will be deployed. Our approach aims to significantly reduce analysis time while maintaining or improving the accuracy of bleeding detection, potentially enabling more timely interventions and better patient care in gastrointestinal medicine.

\section{Methods}\label{sec2}

\subsection{Dataset Development}
In our initial experiments with the MISAHUB challenge dataset, which provided 2,618 wireless capsule endoscopy (WCE) frames containing both bleeding and non-bleeding cases, data quality and quantity emerged as the primary bottleneck for model performance. Our early models showed clear signs of overfitting when trained solely on this initial dataset, indicating the need for a more comprehensive data collection strategy\cite{kvasir}.

To address these limitations, we implemented an extensive data expansion pipeline. We systematically searched for and incorporated additional WCE images from other publicly available medical datasets. Furthermore, we carefully extracted and annotated relevant frames from endoscopy videos to capture a wider range of bleeding presentations. Through this comprehensive data collection effort, we successfully expanded our dataset from the initial 2,618 frames to a total of 6,345 images, significantly enhancing the diversity and robustness of our training data.

The annotation process proved to be equally critical to model performance. Our initial analysis revealed inconsistencies and gaps in the existing annotations, prompting us to undertake a complete reannotation of the entire dataset. This process involved multiple iterations of refinement, with each version showing incremental improvements in model performance. We implemented strict quality control measures during annotation, ensuring that each bleeding region was accurately delineated and properly labeled.

To ensure robust model evaluation, we maintained a careful balance between training and validation data, implementing an 80/20 split ratio. This split was strategically designed to ensure that the validation set represented the full spectrum of bleeding presentations and image variations present in our expanded dataset. The validation set served as a crucial tool for assessing model generalization and preventing overfitting during the training process.

The impact of our comprehensive data strategy became evident in subsequent model iterations. Each refinement in the dataset and its annotations led to measurable improvements in model performance, demonstrating the critical importance of high-quality, diverse training data in medical image analysis applications.

\subsection{Model Architecture}

\begin{figure}[htbp]
    \centering
    \includegraphics[width=\linewidth]{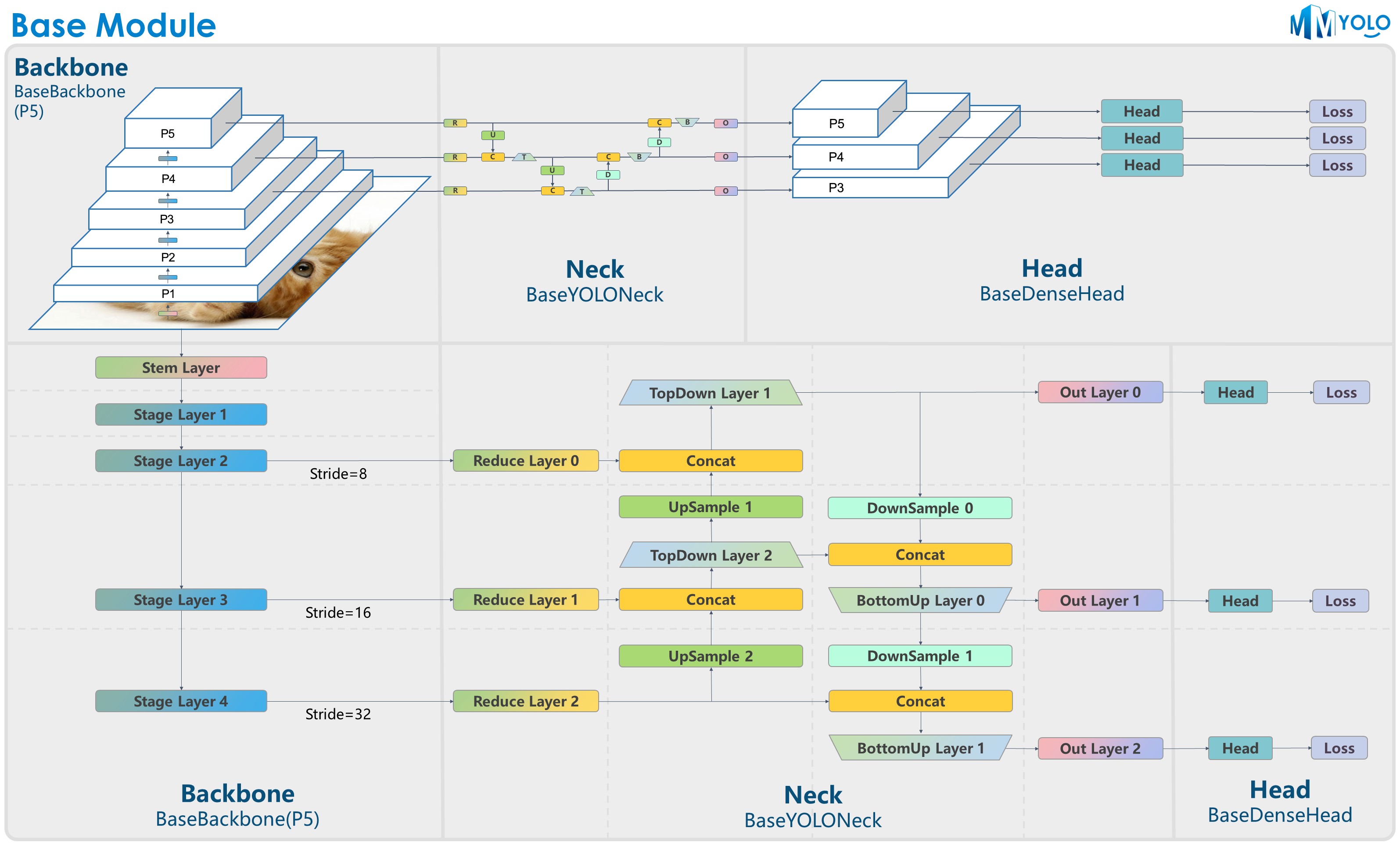}
    \caption{Architecture of YOLOv8-X showing the Backbone for feature extraction, Neck for feature aggregation, and Head for detection outputs, with multi-scale processing using different stride lengths.}
    \label{fig:yolo_architecture}
\end{figure}

Our model selection strategy involved extensive experimentation with multiple state-of-the-art architectures to find the optimal solution for WCE bleeding detection and classification. We conducted our evaluation in two parallel tracks: object detection and image classification. For object detection, we systematically tested the YOLO (You Only Look Once)\cite{yolov8_2024} family, particularly the YOLOv8 variants ranging from small (S) to extra-large (X), along with alternative architectures like RTMDET\cite{rtmdet} and DETR\cite{detr}. For classification, we implemented ResNet\cite{resnet}, EfficientNet\cite{efficientnet}, and a MATLAB-based MobileNet model\cite{mobilenet}. This comprehensive approach allowed us to explore various architectural combinations and their trade-offs between computational efficiency and detection accuracy.

Through extensive experimentation, the YOLOv8-X architecture emerged as the superior performer, demonstrating the best balance of accuracy and generalization capabilities. While smaller variants like YOLOv8-S and YOLOv8-M offered computational efficiency, they showed limitations in detecting subtle bleeding regions. The large (L) variant showed promising results but was ultimately outperformed by YOLOv8-X in terms of detection accuracy and feature extraction capability. Alternative architectures like RTMDET and DETR, despite their theoretical advantages, did not match the overall performance and reliability of YOLOv8-X in our specific medical imaging context.

As shown in Figure \ref{fig:yolo_architecture}, YOLOv8-X's success can be attributed to several key architectural advantages. Its enhanced backbone network demonstrated superior feature extraction capabilities, particularly crucial for identifying subtle bleeding patterns in WCE images. The model's deeper architecture allowed for more nuanced feature learning, while its advanced detection head provided precise localization of bleeding regions. Furthermore, YOLOv8-X maintained real-time processing capabilities despite its larger size, making it suitable for practical clinical applications.

A particularly valuable aspect of our implementation was the unified approach to both detection and classification tasks, though this decision emerged from extensive experimentation with separate specialized models. Initially, we explored a two-model approach, utilizing ResNet and EfficientNet architectures for classification while employing YOLOv8-X for detection. However, this approach revealed significant limitations. The classification models, despite achieving impressive accuracy rates exceeding 99\% on training and validation sets, demonstrated severe overfitting. When evaluated on external datasets, their performance degraded dramatically to around 60\% accuracy, indicating poor generalization capabilities.

These generalization issues led us to reconsider our approach. YOLOv8-X enabled us to handle both tasks within a single framework, offering more consistent performance across diverse datasets. By unifying detection and classification, we not only simplified the deployment pipeline and reduced computational overhead but also achieved better generalization. The unified model proved more robust in real-world scenarios, maintaining reliable performance across varied WCE images. This approach's success in balancing accuracy with generalization made it particularly suitable for clinical workflows, where consistent performance across diverse patient cases is crucial.

\subsection{Training Strategy}
Our training strategy evolved through multiple iterations, closely intertwined with continuous dataset improvements. We began with a foundation of COCO pre-trained weights to leverage transfer learning benefits, but quickly realized that the quality and quantity of our domain-specific data would be crucial for optimal performance.

The training process involved extensive experimentation with various hyperparameters. We systematically explored different combinations of epochs, batch sizes, and learning rates to find the optimal configuration. Through this process, we observed that batch sizes of 16 and 32 provided the best balance between memory usage and model convergence. Learning rate scheduling proved particularly important, with a cosine decay strategy showing better convergence compared to step-based decay.

A key insight from our experiments was the direct correlation between annotation quality and model performance. As we refined our annotations and expanded our dataset, we observed consistent improvements in both detection accuracy and classification reliability. This observation led us to adopt a parallel development approach: continuously improving our dataset while training new model versions. Each iteration of data enhancement was followed by a new training cycle, allowing us to quantitatively measure the impact of our data improvements.

We utilized the Ultralytics framework for implementation, which provided robust tools for training and evaluation. The framework's efficient data loading and augmentation capabilities helped us manage our growing dataset effectively. We implemented a comprehensive monitoring system to track various performance metrics across training iterations, allowing us to identify and address performance bottlenecks quickly.

The final model emerged after numerous training cycles, each benefiting from incremental improvements in data quality and quantity. This iterative approach to training, combined with our parallel dataset enhancement strategy, proved crucial in developing a model that could generalize well across diverse WCE imaging conditions.

\section{Results}\label{sec3}

Our unified YOLOv8-X model demonstrated robust performance across both classification and detection tasks, with particular emphasis on minimizing false negatives given their critical importance in medical diagnosis. We evaluated the model's performance through comprehensive testing on our validation dataset, focusing on metrics that reflect clinical safety and reliability.

\subsection{Performance Metrics}
To provide a comprehensive view of our model's capabilities, Table \ref{tab:metrics} presents the complete set of evaluation metrics achieved on the validation dataset.

\begin{table*}[htbp]
\renewcommand{\arraystretch}{1.4} 
\centering
\begin{tabular}{|p{3.5cm}|p{7cm}|p{2.5cm}|}
\hline
\textbf{Task} & \textbf{Metric} & \multicolumn{1}{c|}{\textbf{Value}} \\ \hline
Classification & Accuracy & \multicolumn{1}{c|}{96.10\%} \\ \cline{1-3}
Classification & Recall & \multicolumn{1}{c|}{96.10\%} \\ \cline{1-3}
Classification & F1-Score & \multicolumn{1}{c|}{96.10\%} \\ \hline
Detection & Mean Average Precision (IoU@0.5) & \multicolumn{1}{c|}{76.8\%} \\ \cline{1-3}
Detection & Average Precision & \multicolumn{1}{c|}{76.8\%} \\ \cline{1-3}
Detection & Intersection over Union (IoU) & \multicolumn{1}{c|}{80.75\%} \\ \hline
\end{tabular}
\vspace{2mm}
\caption{Evaluation metrics achieved by our YOLOv8-X model on the validation dataset. The metrics demonstrate strong performance in both classification and detection tasks, with particularly high accuracy in classification and robust localization in detection.}
\label{tab:metrics}
\end{table*}

The model achieved consistent performance across classification metrics, with 96.10\% for accuracy, recall, and F1-score, indicating balanced and reliable bleeding detection. The identical values across these metrics suggest the model maintains equal sensitivity and precision, crucial for clinical applications. For detection tasks, the model achieved a mean Average Precision of 76.8\% at 0.5 IoU threshold, with an overall averaged IoU of 80.75\%, demonstrating precise localization of bleeding regions.

\begin{figure}[htbp]
    \centering
    \includegraphics[width=\linewidth]{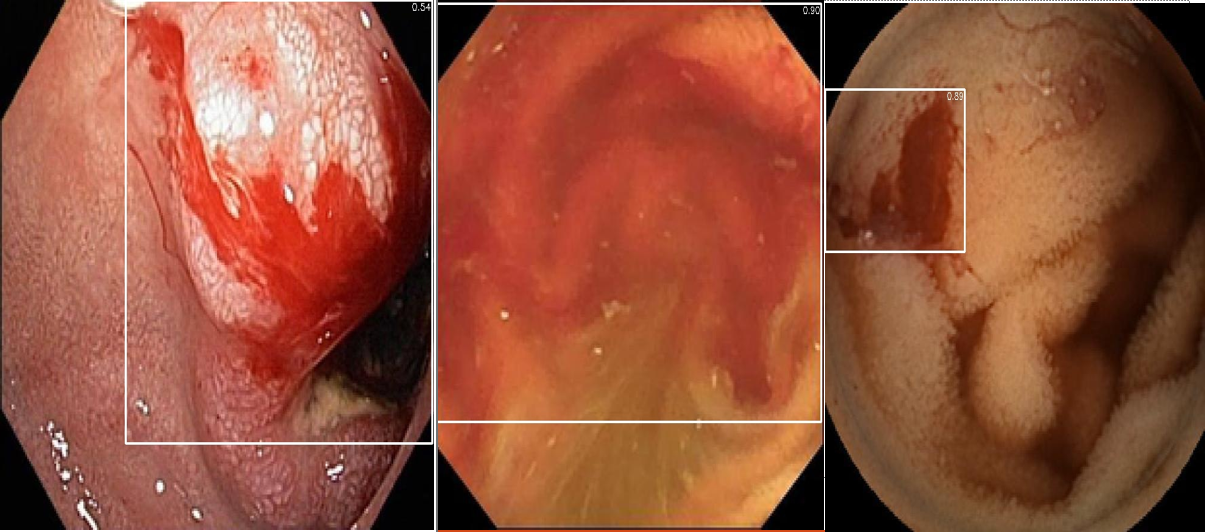}
    \caption{Examples of successful bleeding detection from our validation dataset, showing the model's ability to identify different types of GI bleeding.}
    \label{fig:validation_results}
\end{figure}

\begin{figure}[htbp]
    \centering
    \includegraphics[width=\linewidth]{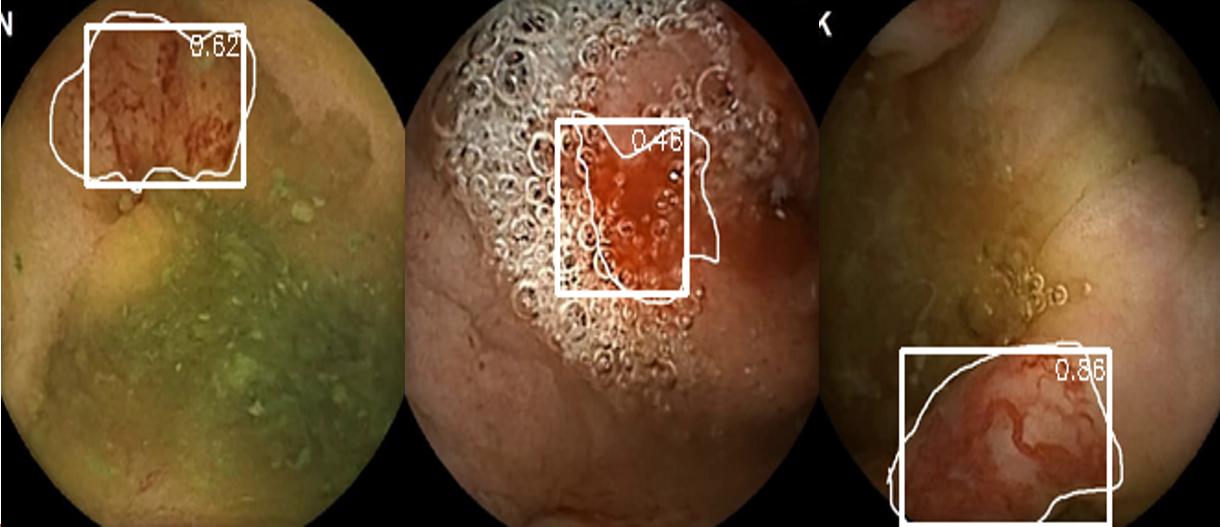}
    \caption{Detection results on Test Dataset 1 showing three different bleeding cases with confidence scores. The white bounding boxes indicate detected bleeding regions with scores of 0.62, 0.48, and 0.88 respectively, demonstrating accurate detection across varying bleeding patterns and lighting conditions.}
    \label{fig:test_dataset_1_results}
\end{figure}

\begin{figure}[htbp]
    \centering
    \includegraphics[width=\linewidth]{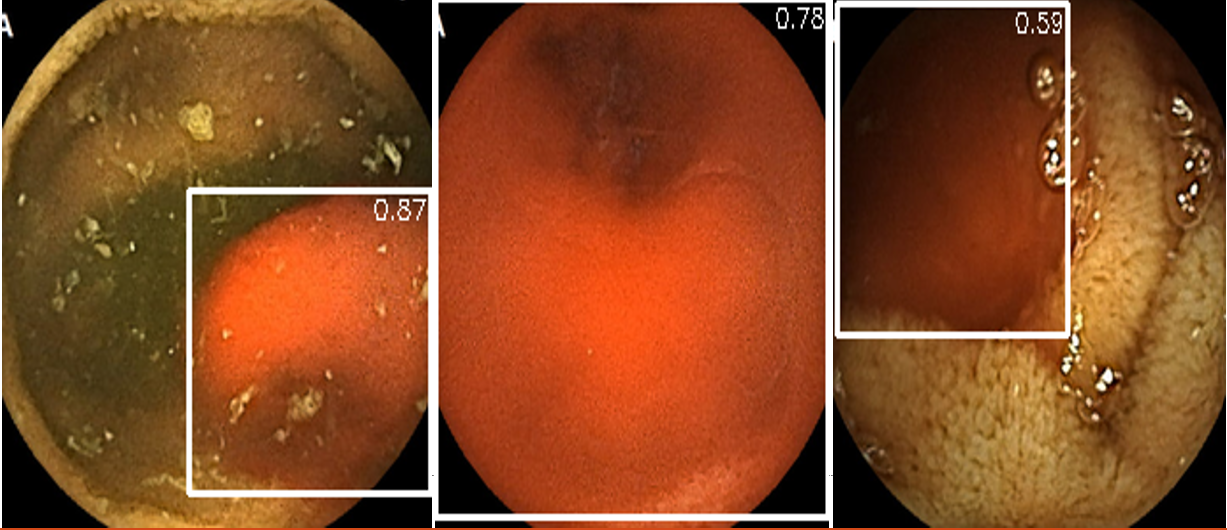}
    \caption{Detection results on Test Dataset 2 showing bleeding regions (marked by white bounding boxes) with confidence scores of 0.87 (left), 0.78 (middle), and 0.59 (right).}
    \label{fig:test_dataset_2_results}
\end{figure}

\subsection{Classification Performance}
The model achieved strong classification metrics that reflect our priority of minimizing missed bleeding cases. The high recall rate of 96.10\% is particularly significant, as it indicates the model's strong capability to identify true bleeding cases, minimizing potentially dangerous false negatives. This was achieved while maintaining an overall accuracy of 96.10\%, demonstrating that this sensitivity to bleeding cases didn't come at the cost of excessive false positives. The balanced F1-Score of 96.10\% confirms that we successfully achieved our goal of prioritizing the detection of all bleeding instances while maintaining clinical practicality.

\subsection{Detection Performance}
In localizing bleeding regions, our model achieved a Mean Average Precision (mAP@0.5) of 76.8\%, with a matching Average Precision of 76.8\%. The Intersection over Union (IoU) of 80.75\% demonstrates precise localization of bleeding regions. We intentionally tuned our detection thresholds to favor higher sensitivity, ensuring that subtle or small bleeding regions weren't missed, even if this occasionally resulted in false positive detections. This approach aligns with clinical priorities where missing a bleeding region (false negative) could have more serious consequences than marking a non-bleeding region for review (false positive).

\subsection{Clinical Relevance}
The model's performance translates into meaningful clinical benefits while maintaining safety:
\begin{itemize}
    \item \textbf{Safety-First Approach:} Our model's high recall rate directly supports clinical safety by minimizing missed bleeding cases. While this may result in some false positives requiring physician review, it ensures potentially critical cases aren't overlooked.
    \item \textbf{Efficient Review Process:} The model reduces analysis time from hours to minutes while maintaining high sensitivity to bleeding regions. Even with occasional false positives, the overall time savings is substantial, and the system ensures critical cases are flagged for review.
    \item \textbf{Consistent Performance:} Unlike human operators who may experience fatigue-related degradation in performance, our model maintains consistent sensitivity across long sessions of image analysis. This reliability is particularly important for maintaining high detection rates across large sets of WCE frames.
    \item \textbf{Diverse Case Handling:} The model demonstrated reliable detection across various bleeding presentations, from obvious to subtle cases. This robustness is crucial for real-world clinical applications where bleeding patterns can vary significantly.
\end{itemize}

Our validation testing included analysis of error patterns, which confirmed that our model's bias toward minimizing false negatives aligns with clinical priorities. While this occasionally results in false positives, particularly in cases of intense regular tissue redness, clinicians have indicated that this trade-off is appropriate for the intended use case of WCE image analysis, where missed bleeding regions could have serious consequences for patient care.

\section{Discussion}\label{sec4}

Our development of an automated bleeding detection system for WCE images provided valuable insights into applying deep learning to medical imaging challenges. Through our experimentation process, we identified key factors that contributed to our success while also uncovering areas for future improvement.

\subsection{Critical Success Factors}
Data quality emerged as the primary driver of our model's performance. Early in development, we discovered that model performance was constrained more by data limitations than by model architecture. We found that carefully curating and annotating our dataset had a greater impact on performance than implementing more complex models. Each time we improved our annotations or added well-labeled data, the model's performance increased notably.

Our decision to use a unified model architecture also proved crucial to our success. While we initially tried using separate models for classification and detection, combining both tasks into a single YOLOv8-X model yielded better results. This unified approach not only simplified deployment but also showed better generalization across different types of WCE images. The model performed consistently well across various bleeding presentations, from subtle to obvious cases.

\subsection{Limitations and Future Directions}
Despite our positive results, we identified several areas for improvement. The main challenge remains data availability. While our dataset of 6,345 images is substantial, medical applications typically require even larger datasets for robust performance. Getting high-quality annotations for medical images is both time-consuming and expensive, creating a significant bottleneck.

Looking ahead, we see several promising directions for improvement:
\begin{itemize}
    \item Collaborating with medical institutions could help us access larger datasets and expert knowledge. This would be particularly valuable for improving the model's performance on rare or unusual bleeding cases.
    \item We're also interested in exploring segmentation approaches. While our current detection-based model works well, precise segmentation of bleeding regions could provide doctors with more detailed information for diagnosis.
    \item Additionally, we could improve accuracy by analyzing video sequences rather than individual frames. Our current approach examines each frame independently, but considering how bleeding patterns develop across consecutive frames could help reduce false positives.
\end{itemize}

As deep learning technology continues to advance and more medical data becomes available, we expect to see further improvements in automated bleeding detection systems. The lessons learned from this project could also help develop similar tools for other types of medical imaging.

\section{Conclusion}\label{sec5}

Our solution to the Auto-WCEBleedGen Challenge demonstrates the significant potential of deep learning in medical image analysis. Through careful attention to data quality and architectural choices, our unified YOLOv8-X model achieved strong performance in both bleeding detection and classification tasks. The model's ability to maintain high accuracy while prioritizing the reduction of false negatives makes it particularly valuable for clinical applications. Our experience highlights the importance of balancing technical sophistication with practical clinical requirements in medical AI development. As we look ahead, our focus will remain on expanding datasets, refining model performance, and strengthening collaborations with medical professionals to enhance the real-world impact of this technology.

\section{Acknowledgments}\label{sec6}
As participants in the Auto-WCEBleedGen Version V1 Challenge, we fully comply with the competition rules as outlined in \cite{hub2024auto} and the challenge website. Our methods have used the training and test data sets provided in the official release in \cite{palakbleedingtrain} and \cite{palakbleedingtest} to report the results of the challenge.

We thank CVIP 2023 and IIT Jammu for organizing this challenge and advancing medical image analysis research. Special thanks to Dr. Palak Handa, Dr. Deepti Chhabra, and the organizing committee for their excellent management of the competition.

We are grateful to IIT Indore and the Pattern Recognition and Image Analysis (PRIA) Lab for providing the resources and environment for this research.

We also thank the broader medical imaging community whose research continues to advance automated medical diagnosis.

\bibliographystyle{plainnat}
\bibliography{references}  

\end{document}